# Graceful Forgetting
# II. Data as a Process

Alain de Cheveigné
Centre National de la Recherche Scientifique
Ecole normale supérieure, Paris

This is the second part of a two-part essay on *memory* and its inseparable nemesis, *forgetting*. It looks at memory from a computational perspective in terms of function and constraints, in the rational spirit of Marr (1982) or Anderson (1989). The core question is: How to fit an infinite past into finite storage? The requirements and benefits of such a "scalable" data store are analyzed and the consequences explored, the main of which is that conserving data should be seen as a *process*, possibly complex to design and expensive to compute, but nonetheless worthwhile. The first part of this essay (*Memory as a Process*) discusses how these ideas translate – or not – into insight about memory in humans or other biological systems.

## Abstract
Data are rapidly growing in size and importance for society, a trend motivated by their enabling power. The accumulation of new data, sustained by progress in technology, leads to a boundless expansion of stored data, in some cases with an exponential increase in the accrual rate itself. Massive data are hard to process, transmit, store, and exploit, and it is particularly hard to keep abreast of the data store *as a whole*. This paper distinguishes three phases in the life of data: *acquisition*, *curation*, and *exploitation*. Each involves a distinct process, that may be separated from the others in time, with a different set of priorities. The function of the second phase, curation, is to maximize the future value of the data given limited storage. I argue that this requires that (a) the data take the form of summary statistics and (b) these statistics follow an endless process of rescaling. The summary may be more compact than the original data, but its data structure is more complex and it requires an on-going computational process that is much more sophisticated than mere storage. Rescaling results in *dimensionality reduction* that may be beneficial for learning, but that must be carefully controlled to preserve relevance. Rescaling may be tuned based on feedback from usage, with the proviso that our memory of the past serves the future, the needs of which are not fully known.

## Data and statistics

Data are central to science and society. Expressions such as "*big data*", "*data deluge*", "*data are the new oil*" convey the awe – and misgivings – inspired by their exponentially increasing size and domineering role in economy and society. Hate them or love them, big data are here to stay, for science, technology, and society at large. The Britannica Dictionary defines "*data*" as "*facts or information used usually to calculate, analyze, or plan something*", emphasizing their role to support future understanding and action. Wikipedia adds that they are "*often numeric*", implying a contrario that occasionally they might not be numeric. I'll argue that, in the limit of large data, *all* data are numeric and consist of statistics.

A statistic is defined as "*any quantity computed from values in a sample*" (Wikipedia), i.e. a number informative about multiple items of data. Examples are the *cardinality* of a dataset, or the *mean*, *variance*, *min* or *max* of a set of numbers, or a *subset* of data items. Slightly confusingly, the same word designates the operation and the resulting value, and in the plural, *statistics* might refer to several such operations or values, or to statistical inferences based on the data (e.g. significance tests), or to the science or art of applying these tools to make sense of data. Which meaning is intended is usually clear from context.

When data are replaced by a statistic, *information is lost*, the upside being a summary representation that is cheaper to store. For applications (e.g. machine learning) that use statistics as a first step in their calculations, the summary may be as good as the data. The summary can be understood as a parametrization of the *empirical distribution* of the values that it summarizes, enabling inferences to be made about the data, or about the process that produced them, in the absence of the data themselves. For this to work, the values of the statistic must be associated with a *label* or descriptor that documents how they were derived from the data.

It is common to distinguish between raw data and summary, but we can do away with this distinction. In many cases, the observation is already a form of statistic derived from unobserved physical quantities. A sensor might produce a signal that summarizes multiple microscopic quantities, as when a microphone measures acoustic pressure (Fig. 1), and instantaneous values of that signal might be further summarized over time by convolution with an anti-aliasing filter, before being sampled and converted to a digital representation. In this sense, each raw observation is already a statistic of underlying quantities being probed. Note that from now on "sample" is used in the signal-processing sense of "datum", rather than in the statistics sense of subset of a wider population.

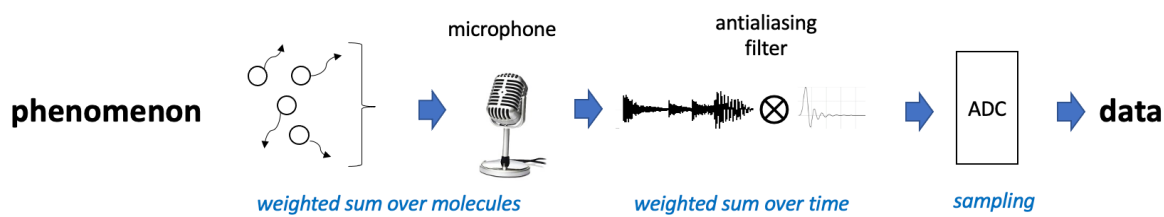

Fig. 1. Raw data as statistics over microscopic quantities. The voltage recorded from a microphone summarizes the motion of multiple molecules, its instantaneous values are convolved by the impulse response of the antialiasing filter, before sampling by the analog-to-digital converter. Each data sample (datum) is thus a statistic over the many microscopic and instantaneous values that characterize the phenomenon being observed.

While some data might initially be symbolic, categorical, or structural, in the limit of very large data what counts is the *number* of each symbol, category or structure, or statistics of the distribution of that number. Since we are mainly interested in that limit, there is little point in singling out initial observations as special. This justifies, as a working hypothesis:

> *Idea 1: In the limit of large data, all data are statistics[1]*

## Three phases in the life of data

Data go through three phases: *acquisition*, *curation,* and *exploitation* (Fig. 2). The second phase, curation, is usually thought of as merely passive storage but I'll argue that it should be seen as process, possibly more sophisticated than the other two phases.

The three phases are sometimes concurrent, as in the closed-loop control of an industrial process (grey arrow in Fig. 2). In other cases, the phases are well separated in time, possibly over a long period, and possibly performed by different *actors*, under different *constraints*, and serving different *goals*. Data might be gathered by an experimentalist or a data-logging system, curated by a system manager or librarian, and exploited by a data analyst or theoretician. Data gleaned by an app (e.g. "likes" or accelerometer readings) might be repackaged and sold for marketing purposes, or used by a different app for an unforeseen purpose. This paper is mainly interested in the accumulation of data for such future, unforeseen uses.

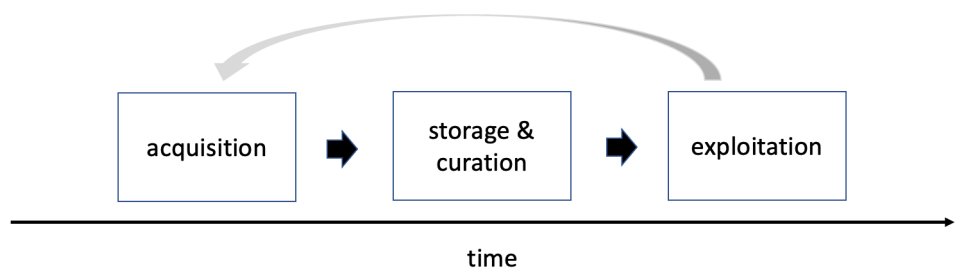

Fig. 2. Three phases in the life of data. Data are acquired by one process, conserved by a second, and exploited by a third. The three phases might be concurrent within a tight loop (grey arrow), or instead well demarcated and separated over a long time-span.

> *Idea 2: Data acquisition, conservation, and exploitation are distinct processes, with distinct objectives and constraints.*

The distinction between phases is important because the first two phases (acquisition and curation) must make decisions that affect the usefulness of the data for the third (exploitation). Causality implies that the value of the data, revealed in the third phase, is unknown to the first two, so decisions such as "what should I record?" or "what should I conserve?" are blind to that value. The value can be *guessed*, for example based on past exploitation of similar data, but the guess is uncertain, particularly if the phases are separated by a long period. Acquiring data is like putting a message in a bottle without knowing who will read it.

---
[1] A series of "Ideas" are singled out to mark steps in the argument. Some are straightforward, others less, with no claim of novelty. See Discussion for pointers to similar ideas in the literature.

> *Idea 3. Acquisition and conservation have incomplete knowledge of the value of the data.*

Acquisition and curation face difficult design decisions (e.g. sampling rate, number of bits, etc.). This paper hints at how to make those decisions easier.

## Conservation as an active process

Data storage is usually thought of as passive. However, if new data arrive at a steady rate, sooner or later the storage will be full (unless technological progress allows it to grow faster than the data, an uncertain assumption). The system is then faced with a hard decision: refuse new data or delete old data.

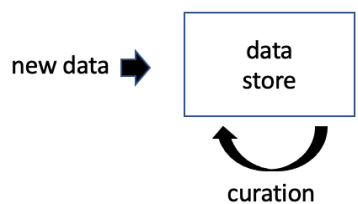

Fig.3. Data conservation as an active curation process. If input is unlimited and storage is limited, old data must be *deleted*, or new data *ignored*, or stored data made *more compact*. This occurs repeatedly, implying an ongoing process, active throughout the life of the data.

Refusing or deleting, voluntarily or not, is a form of statistic according to our definition, and it is logical to consider other statistics that might serve the same purpose with a more useful outcome. *Curation* consists in selecting a statistic, choosing its parameters, and carrying out the calculations to create or maintain a summary of the data.

> *Idea 4: Data conservation involves an active process of curation.*

The system that runs the curation process needs to know what to do in every situation: it must have instructions (or a program) indicating the *rules* or *algorithms* to apply. If they are data-specific, it is convenient to store them with the data.

> *Idea 5: Data must be augmented with instructions as to how to curate them.*

This is all the more important as the data might be transferred between systems (e.g. computers or cloud services), repeatedly during their lifetime. Each system needs to know what to do with the data when it runs out of space. The instructions too can evolve as part of the curation process. A default set of instructions might be defined based on an educated guess of future needs, and later revised based on new insights gleaned from the world.

> *Idea 6: Instructions stored with the data may be curated together with the data.*

Each step of data curation implies *loss of information*. The loss is irreversible, so the choice of which part to discard should be made wisely. As time goes on, the relative value of different parts of the data can better be estimated, and therefore the decision to rescale the statistics should be made *as late as possible*. Curation operations should be triggered only when necessary to make space for new data.

> *Idea 7: Data curation should be lazy. Its operations should be deferred to the latest possible time, triggered only to avoid uncontrolled data loss.*

Data curation might be justified for other reasons, for example to remove irrelevant or untrustworthy information, or reduce dimensionality, or ease search, etc. However, curation for those purposes can be deferred as long as storage is available: lack of storage is the only compelling reason to discard information now[2]. Links between summarization and search and dimensionality reduction are discussed further on.

To summarize, the picture emerges of a curation process that is active over the entire lifetime of the data. The process might be simple, guided by hardwired rules, or sophisticated, guided by a program that is refined over time based on newly available information. The summary should be augmented by (a) information that documents how it was derived so that inferences can be made about the original data, and (b) instructions that specify how it should be treated when space runs out, to guide the curation process. Curation should be lazy, to capitalize on the possibility that better rules can be found based on newly gleaned information.

The greater complexity and processing requirements, relative to passive storage, are justified, or not, depending on the future value of the data, with the caveat that the true value is unknown and can only be wagered. A complex process may be worthwhile, despite complexity, if the potential value is large.

## Basic statistics

Replacing $N$ numbers by their *mean* summarizes the data in a form that is $N$ times more concise. The mean indicates the central tendency, augmenting it with the *variance* (or *covariance*) indicates the spread. Both can be used as parameters of a Gaussian model of the empirical distribution of values within the original dataset. Higher-order moments (e.g. *skewness* and *curtosis*) quantify deviations from a Gaussian distribution, extrema (*min* and *max*, or *convex hull*) provide strict bounds, the *histogram* can describe an irregularly-shaped empirical distribution, and so-on. Other useful statistics are described in Appendix A.

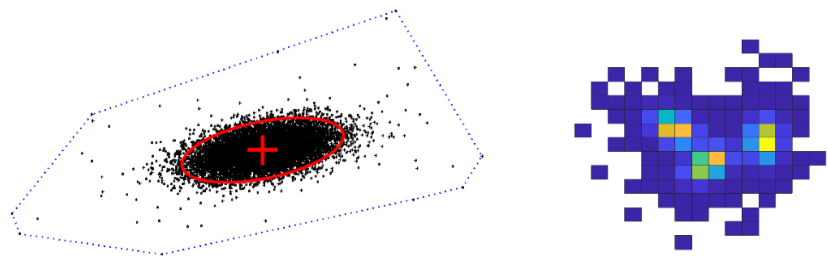

Fig. 4. Left: a cloud of 2D points, of size $N = 10000$, is summarized by their mean (red cross), covariance (symbolized by a red ellipse), and convex hull (apices of blue polygon). Right: a cloud of 2D points is summarized by a histogram. The representation on the left requires 21 floating point numbers, that on the right about 110 integers (not counting the rule or dictionary that defines the bins).

---

[2] For simplicity, computational cost is ignored.

*Idea 8: Statistics parametrize a statistical model of our knowledge (and uncertainty) about the data they summarize.*

If new data are acquired continuously, the statistics need to be applied repeatedly. The result is a *time-series* of statistics, each sample summarizing an interval of the original data.

*Idea 9: If data input is continuous, the summary forms a time series.*

This time series constitutes a statistical representation of the original data. Information about the temporal order of samples *within* each interval is lost, but order *between* intervals is preserved. If we don't care about time, the summary can also be interpreted as parameters of a mixture model (e.g. mixture of Gaussians) of the distribution of the original data considered as a bag of values. The richer the summary (more samples of the statistics), the more detailed the statistical model.

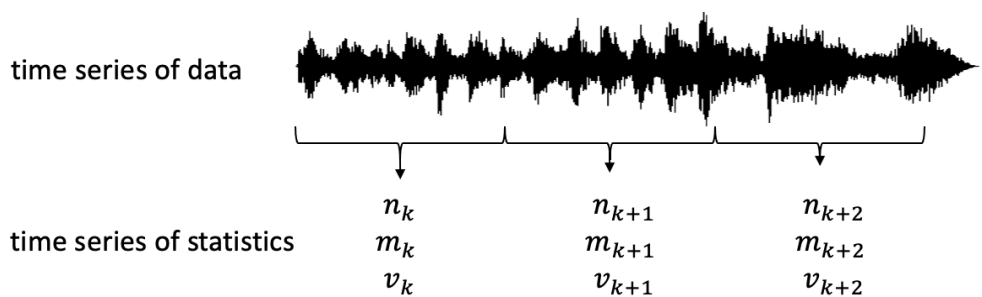

Fig. 5. Summarizing data as a time series of statistics. Here, each interval is summarized by its cardinality, mean, and variance. Statistics might also include bounds (min, max), a histogram, a power spectrum, or more complex statistics as described below.

## Scalability

As new samples of the summary statistics are appended to the old, the summary grows without bound. We are confronted with the same problem as for the raw data, that can be addressed by forming a *statistic of statistics* which is, by definition, a statistic of the original data. One can ask under what conditions the result is the same as if that statistic had been calculated directly, a property referred to here as *scalability*.

The sum is scalable: the sum of sums $s(A)$ and $s(B)$ of two disjoint subsets is equal to the sum of their union, $s(A \cup B)$. Cardinality is scalable since we have $n(A \cup B) = n(A) + n(B)$ for $A$ and $B$ disjoint. As a counter-example, the mean by itself is not scalable unless it is associated with cardinality: the mean of the union of disjoint intervals is their cardinality-weighted mean: $m(A \cup B) = [m(A)n(A) + m(B)n(B)]/n(A \cup B)$. Variance associated with mean and cardinality is also scalable, as are extrema, histogram, and so on. As another counter-example, the median is not scalable since the median of the union of A and B is not a function of their medians.

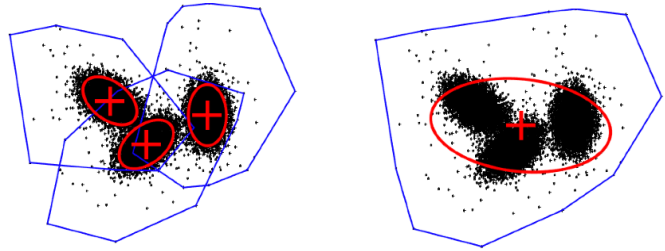

Fig. 6. Scalable statistics. Left: three clouds of 2D points, each of size $N = 10000$, are summarized by their mean (red crosses), covariance (symbolized by a red ellipse), and convex hull (apices of blue polygons). Right: the representation is rescaled to a single mean, covariance matrix, and convex hull, a cruder but less costly summary. The same data were represented as a histogram in Fig. 4 (right).

Scalability is useful because it avoids having to document the exact sequence of calculations that produced the statistics.

*Idea 10: Data should be summarized using scalable statistics.*

## The arrow of time

As new data come in, rescaling must be performed *repeatedly*. Each rescaling entails information loss, and thus older portions of the original data stream are represented with fewer bits than newer portions. The older a datum, the more rescaling it will have undergone.

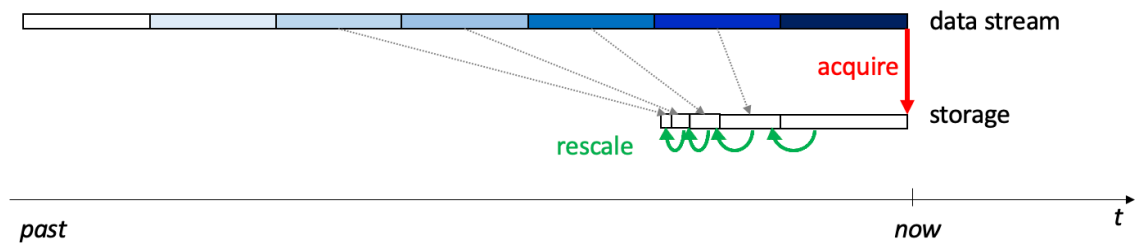

Fig. 7. Rescaling automatically discounts old information relative to new. As a new chunk of the data stream (top) is acquired and stored, each of the previously stored interval is rescaled to take less room. Ancient data have been rescaled more times than recent (as symbolized by a paler colour).

This neatly matches the empirical observation that *older data elicit less interest than new data*, all else being equal. Citations to older papers are less likely than to recent ones (Golosowsky and Solomon 2016), the World Wide Web shows less interest for events remote in time (Jovani and Fortuna, 2007), and similar effects are seen in libraries (Burrell, 1985), and online attention (Candia et al., 2019). The trend is congruent with Fitts's law which, applied to time, would suggest that the uncertainty with which we locate a datum in time is proportional to its age[3]. It is also consistent with Bialek et al's (2001) theoretical argument that predictive information grows sub-extensively. Each new fact grabs some attention, and thus the share of previously acquired facts dwindles over time.

---

[3] Fitts's (1954) law says that the time required by a subject to point to a target is a function of the ratio between the distance to target and the target width. The law applies to many situations.

> *Idea 11: If an unlimited data stream is stored in a finite store, older data tend to be assigned less importance than new.*

In addition to these dilution effects, old facts may be *less relevant in a changing world*. However, they may still carry important information, all the more so as there are *lots* of them, and this justifies the effort to keep a trace of even the oldest data. Relevance depends on factors other than time.

As mentioned earlier, the time series of statistics preserves slow fluctuations in the data, but faster fluctuations are lost, *unless captured by a statistic*. One such statistic is the short-term *power spectrum* (Appendix A) which, with an appropriate definition, is scalable and more concise than the raw data. The spectrum summarizes the *pattern of fluctuation* of the data at various time scales. For data remote in time, a spectral summary may be more useful than a record of individual values.

Applied to consecutive intervals, the spectrum statistic results in a time series of vectors of spectral coefficients. As time proceeds further, it may be useful, and necessary, to concisely summarize them as a *second-order power spectrum* (also called *modulation spectrum* in psychoacoustics and speech processing), which itself can be summarized as third-order spectrum, and so on. Such a hierarchical analysis scheme has been formalized as the *scattering transform* (Anden and Mallat, 2014; Bruna and Mallat, 2013) or *modulation cascade processes* (Turner and Sahani, 2008). With appropriate definitions, all of these statistics are scalable.

Simple statistics (mean, etc.) can be understood as parameters of the distribution of raw data samples, and as such are informative about *individual values*. A spectral statistic, particularly higher-order, is best understood as informative about an *interval of values*, reflecting the "texture" of the data within that interval (defined circularly as an abstract property of the data that results in a particular statistical description). A statistic can thus play two roles: parameter of a distribution, or descriptor of an abstract property. In the limit of large data, the latter is likely to be more important.

> *Idea 12: Statistics are informative as descriptors of abstract properties (texture) of intervals of data, useful in the limit of large data.*

Appendix A provides other examples of scalable statistics. A variety of statistics is available, which is both a blessing, in that it provides flexibility to optimize the summary according to the nature of the data and the needs of an application, and a curse in that there may be uncertainty or disagreement as to which statistic(s) to choose. The same data might be summarized by different statistics, masking the fact that they are the same. Curation and exploitation processes must be able to abstract over those differences so as to be able to treat the summary uniformly as a memory of the data.

> *Idea 13: The same data might be summarized by different statistics. Curation and exploitation must abstract over those differences.*

## Inference, search, and dataset comparison

A statistical summary is not only necessary, but useful. As parameters of an empirical distribution, statistics allow inferences for example to judge the likelihood that a particular value was among some subset of the original data. This makes them useful to construct an index to search within those data. Efficient search hinges on the ability to *prune* or *prioritize* the search space, which can be done probabilistically (and in some cases deterministically) based on statistics.

Figure 8 illustrates the search for a query value within a space of 2D points (left), using a hierarchical index (tree with statistics at each node). The tree can be constructed from its leaves (e.g. a time series of statistics) thanks to the scalability property. Here, the search concludes rapidly that the query is not in the data, without the need to access the data. If these were a part of a larger search space, this part could rapidly be pruned.

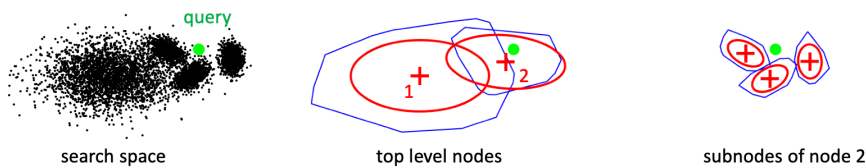

fig. 8. Search aided by a hierarchical tree of statistics. Search for the query point (green) is assisted by a hierarchical index with nodes labelled with statistics (center and right). Based on convex hull statistics, node 1 is ruled out (center), and then each of the subnodes of node 2 (right), concluding rapidly that the query is *not* among the data points.

*Idea 14: A statistical summary supports search.*

Another useful task is to *compare* data sets based on their statistics. Are datasets $A$ and $B$ the same, or is one a subset of the other? A difficulty is that they might have been labelled with different statistics, or by the same statistics but rescaled differently. Thus, the data might be the same but the summaries different. The same issue plagues search for data characterized by their statistics ("texture"): the query token might have been summarized by a different statistic than was applied to the search space. This issue is likely to become serious in the limit of large data.

## Kullback-Liebler divergence

The situation where the same data have been labelled by different statistics can be addressed by using a statistical *divergence* measure to compare distributions that they parametrize. Given distributions $\mathcal{A}$ and $\mathcal{B}$ parametrized by the statistics that summarize datasets $A$ and $B$, the Kullback-Liebler (KL) divergence

$$D_{KL}(\mathcal{A} \parallel \mathcal{B}) = \sum_{x} \mathcal{A}(x) log(\mathcal{A}(x)/\mathcal{B}(x))$$

indicates the degree to which $\mathcal{A}$ fits within $\mathcal{B}$. A small value suggests that distribution $\mathcal{A}$ fits within $\mathcal{B}$, indicating that it is plausible that $A$ is a subset of $B$. The divergence is not symmetrical: $D_{KL}(\mathcal{B} \parallel \mathcal{A})$ might be large, indicating that $B$ is likely not a subset of $A$.

If both $D_{KL}(\mathcal{A} \parallel \mathcal{B})$ and $D_{KL}(\mathcal{B} \parallel \mathcal{A})$ are small (or even zero), the datasets *might* be identical. This can be *corroborated* if more detailed statistics are available, for example with higher temporal resolution, but *confirmation* is possible only if the raw data are available. In the absence of raw data, corroboration is the best we can do.

> *Idea 15: KL divergence can abstract across different possible summarizations, allowing datasets to be compared on the basis of their summaries.*

As a simple example, extrema statistics applied to datasets *A* and *B* can be used to parametrize uniform distributions (a natural choice for these statistics, Fig. 9, left). Values of *x* for which $\mathcal{A}(x)$ is zero and $\mathcal{B}(x)$ is non-zero make infinite contributions to the Kullback-Liebler divergence $D_{KL}(\mathcal{B} \parallel \mathcal{A})$. This implies that *B* is not a subset of *A*, although *A* could still plausibly be a subset of *B*. The same procedure can be applied if the datasets are parametrized by different statistics (Fig. 9, right).

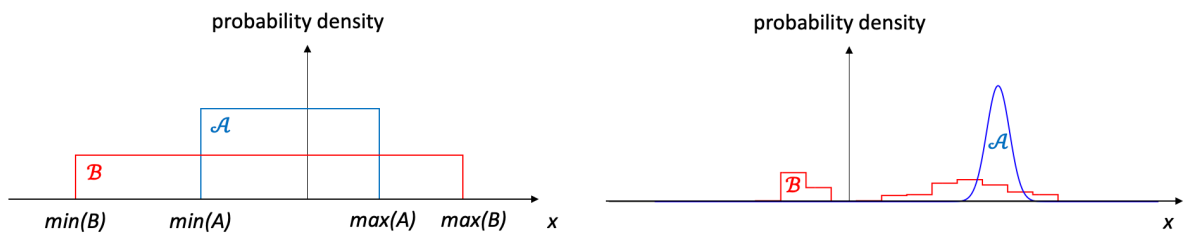

Fig. 9. Left: datasets *A* and *B* are summarized by extrema statistics and modelled as uniform distributions. Right: *A* is summarized by its mean and variance, and *B* is summarized by a histogram, used to parametrize Gaussian and piecewise uniform distributions respectively.

The value of the KL divergence depends also on the choice of parametric family for each statistic (e.g. Gaussian vs uniform), which is arbitrary. It might be useful to augment each statistic with a label suggesting a parametric family, for example determined from the data during the summarization process.

This heuristic and others are discussed further on. Statistical distribution comparison metrics become more important in the limit of large data, where data values become less important than their distribution.

> *Idea 16: In the limit of large data, comparisons are between distributions, rather than between values, requiring tools such as KL divergence.*

## Non-uniform and non-sequential summaries

It was tacitly assumed so far that the statistical summary consists of a time series with uniform (or uniformly graded) sampling resolution. This section makes the point that it may benefit from non-uniform sampling and non-sequential data structures such as strands or episodes. Those structures are discussed in greater detail in Appendix A.

Even if raw data are uniformly sampled over time, they might benefit from a non-uniformly sampled series of statistics (Fig. 10).

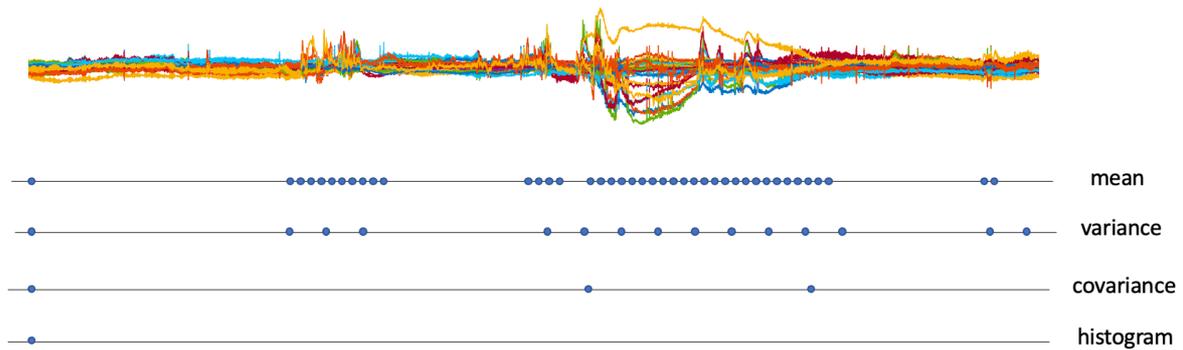

Fig. 10. Top: raw data (multivariate time series). Below: non-uniformly sampled statistics. Cheap statistics (mean, variance) might be instantiated more densely than expensive statistics (covariance, histogram), and with more values assigned to busy than to relatively stationary portions.

If a data stream reflects multiple concurrent processes, it may be useful to parse it into parallel *strands*, each represented by its own time series and rescaled independently from the other strand. For example, we lead in parallel a personal and a professional life (Kubovy 2015). A technique such as Independent Component Analysis (ICA) might split data into components reflecting individual sources, and so on.

Events of interest often *recur*, and indeed their recurrence may be a cue that they are of interest. Recurrent events can be represented concisely by replacing each occurrence by a pointer into a dictionary of *episodes* (as in vector quantization, Gray 1984). Strands and episodes both involve a non-sequential structure.

> *Idea 17: Data are observed in sequence, but their summary may benefit from a non-sequential representation.*

These non-sequential schemes are discussed in greater detail in Appendix A. For the present discussion we retain that a summary may require a data structure that is considerably *more complex* than the original time series, albeit more compact.

> *Idea 18: The statistical summary is likely to be more complex than the data.*

## Curation is a complex process

Curating a complex data structure requires a complex curation algorithm. This is evident if we consider the requirements of, say, an episodic representation (comparison of incoming data with dictionary tokens, curation of the dictionary), and will become even more obvious later on.

> *Idea 19: Curation is likely to be more complex than mere storage.*

Curation of a dictionary for the purpose of a histogram or episodic representation, (Appendix A) involves intensive access to the summary itself, for example to compare new data to existing entries.

> *Idea 20: A primary client for memory is the curation process that reorganizes it.*

It emerges also from the discussion of dictionary curation (Appendix A) that curation may involve a large number temporary entries that are consolidated later, a process all the more effective as there is storage space for these temporary entries. It was also argued (Idea 7) that deferring curation operations may be beneficial, suggesting again that more storage can yield a record of better quality (in addition to more extensive).

> *Idea 21: More storage translates to better memory (in addition to more memory).*

In summary, there are many ways to maintain a statistical summary of the data that is expressive, compact, and scalable.

## Dimensionality reduction

If each observation within the original data is seen as one dimension of a very large observation space[4], summarization amounts to projection (or some other transform) into a lower-dimensional subspace or manifold. The summary lives in a smaller space than the original data.

> *Idea 22: Summarization is a form of dimensionality reduction.*

Dimensionality reduction may be beneficial to avoid overfitting when a model is learned. However, the quality of the model will suffer if relevant dimensions are lost. The choice of statistics (and their parameters) determines which dimensions will be discarded, and which will survive. The wide variety of statistics and parameters available for summarization offers leeway to select dimensions to preserve.

> *Idea 23: The choice of summary statistics, and their parameters, influences which dimensions will survive dimensionality reduction.*

It is important to choose wisely to avoid losing relevant dimensions, keeping in mind that that choice is made during the curation process at a time when the requirements of the application (e.g. learning) are not perfectly known. A number of heuristics can guide this choice.

## Heuristics

One heuristic is implemented automatically by repeated summarization: *recent observations are assigned more dimensions than old*. This is expected to be beneficial, overall, on the assumption that recent observations are more relevant than old (Idea 11). Other heuristics include *slowness* (as exploited by Slow Feature Analysis, SFA, Wiskott and Sejnowski 2002), *nonstationarity* (busy intervals are more relevant than quiescent), *repetition* (a pattern that recurs is unlikely to be random), *prior access* (data judged relevant in the past may be relevant in the future), *correlation* with other data streams, actions or rewards (patterns contemporary with a reward may be relevant), *KL divergence* between subsets of the data (small values suggest that the statistic is indiscriminate), domain-specific *invariances* to transforms such as shift, rotation, or scaling, as exploited by CNNs (convolutional neural networks, LeCun 2015, Fukushima 1980), *joint diagonalization* (a general technique to isolate relevant dimensions) and so on. These heuristics are discussed in more detail in Appendix B.

---

[4] To be clear: a dimension is assigned to each individual observation (datum). A $N$-sample dataset is a point in $N$-dimensional space, the mean projects it into a 1-D space.

*Idea 24: The choice of statistics and parameters may be guided by heuristics.*

Most of these heuristics are not specific to a particular application, and thus they preserve the generality of the summary and its ability to address unknown needs of future applications.

*Idea 25: Heuristics exploit properties based on logical or physical arguments of general relevance, and are only loosely related to specific needs of an application.*

However, different heuristics suggest different priorities, and the choices made by the curation process (and thus the future value of the summary) depend on how these heuristics are ranked or combined. This is a possible target for learning based on feedback from the exploitation phase (more precisely: from the outcome of previous exploitation of similar data).

*Idea 26: Heuristics compete with each other. The priority or balance between heuristics can be tuned for the benefit of a particular application.*

## Generality vs optimality

There is a trade-off between optimizing the summary for a task (via the choice of heuristics) and the need to maintain its usefulness for *tasks unseen* for which that optimization might be counterproductive. As pointed out earlier, the curation phase has imperfect knowledge of the needs of the exploitation phase (Idea 3). The generality of the data record should not be compromised by excessive tuning.

*Idea 27: The urge to tune heuristics for the needs of an application should be resisted, so as to preserve the value of the record for future unseen applications.*

## Actions and rewards as data

Parallel to the observations recorded by the acquisition process, a potentially useful stream of information consists of previous *actions* and consequent *rewards* (positive or negative. Causality dictates that only *past* actions and rewards can be taken into account, in which respect they are analogous to observations. Actions and rewards that occur now might become relevant later on so it is worth conserving them. To save storage they may need to be summarized, and their summary curated, and it makes sense to curate all these streams together in one process as in Fig. 11 (an unrolled version of Fig. 2).

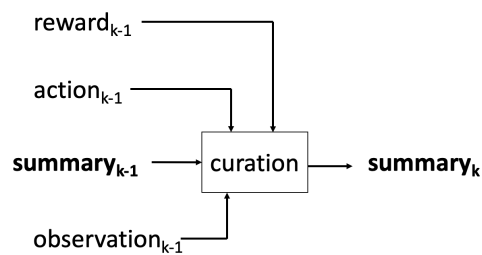

Fig. 11. The previous summary is combined with new observations, actions, and rewards to produce a new summary of all past observations, actions and rewards.

> *Idea 28: Rewards and actions should be summarized together with data and jointly curated.*

## Data as a service

The *prior access* heuristic is only applicable if the curation process knows which parts of the summary are used. One way to ensure that this is the case is for the curation process to offer access to the summary as a *service*. Rather than search within a local copy of the summary, the exploitation process sends a query to the curation process, which in return benefits from knowing which aspects of the summary were of interest. An additional advantage is that the summary does not need to be copied and is always up to date, and the curation process can centralize sophisticated access services (e.g. inference or search).

> *Idea 29: Providing access to the summary as a service allows the curation process to collect usage information to guide future curation.*

## The viewpoint of the exploitation process

The exploitation process inherits a summary record of the past (either the record itself, or the ability to query it) that is an *approximation* of the past. It faces two challenges. One is that it must make do with an approximation, the other is that this approximation depends on parameters of the summary, which it does not control.

The approximation limits the form of function that can be applied to past data. Instead of any function $f(\{x(i)\}), i = -\infty, \ldots, t$ of past observations, the process is limited to functions of the form $f(\{s_j\}), j = 1 \ldots J$ where each $s_j$ is a predetermined function of some subset of the observations. This induces a form of architectural or *inductive bias* (Botvnick et al 2019) on what can be learned from the past. The bias depends on the choice of statistics and parameters, which itself depends on the balance of heuristics. Tweaking this balance on the basis of reward, for example, would implement a form of *slow learning* (Botvnick et al 2019).

## Recap

### The summary

The data record consists of three parts: (1) a data structure containing a statistical summary of the past, (2) a record of how it was derived from the original observations, (3) rules for subsequent curation. Parts (1) and (2) are useful for the exploitation process, all three are required for the curation process.

### Acquisition

The acquisition process must label the data to document how they were obtained, and provide initial rescaling rules and any information useful for curation. In return, it does not need to worry about sampling rate or resolution, etc. because the curation process offers the fiction of infinite storage.

### Curation

The curation process is ongoing, throughout the lifetime of the data. It might see periods of quiescence (passive storage), or even a terminal event such as deletion. In general, however, it remains active to accommodate new data with three priorities: (1) maintain the integrity of the data record, (2) accommodate unlimited incoming data, (3) optimize the data record for future exploitation. Curation involves repeated rescaling to avoid running out of space, condensing old statistics into new statistics. The choice of statistics, parameters, and rescaling rules can be steered on the basis of heuristics, keeping in mind that the purpose of recording the past is to guide *future* needs, which can be guessed but not known.

### Exploitation
The exploitation process is offered an approximation of the past in the form of a statistical representation, but it must be able to *abstract* over variants of this representation that depend on choices made by the curation process. The data record is mined opportunistically for whatever information is of use, with the guarantee that it reflects the original data in a well-defined way. Exploitation can be concurrent with acquisition and curation in a tight loop (grey arrow in Fig. 2), in which case curation choices can be optimized for its needs. In general, however, exploitation occurs after those choices have been made.

## Discussion
### The past as a process
According to the nineteenth-century physiologist Marie Jean Pierre Flourens, "Science *is* not: It *becomes*." (Boring 1963). A message of this paper is that a similar formula applies to a record of the past: *Data are not. They become.* The idea is a bit unsettling: we can accept that the future is uncertain, and the present fleeting, but the past, surely, is firmly established? It is hard to admit that, in a blink of the eye, the past is not the same past as it was before the blink. We expect of data that they be – unless tampered with – immutable. Curation seems akin to tampering, i.e. *wrong*, until of course we realized that tampering *must* occur, be it active by summarizing, or passive by selection or deletion,

The past helps to predict the future and plan for it. All of the past is potentially relevant, but we cannot keep it all because unbounded input can't fit in bounded storage. The question is, *which* information should we let go. That question is asked anew each time new information comes in, and each time an answer must be found and implemented, the process that we call curation. If requirements of the application are foreknown, the process can be tuned possibly via a learning process. If not, choices must be made based on general, application-agnostic grounds. In general, however, there is some precedent in dealing with similar data or needs, and this may allow learning possibly on a very long time-scale (e.g. evolution, Gillings 2016; Botvinick 2019).

Curation emerges as a sophisticated process that is active even when the application is quiescent. It is a prime client for memory itself, and possibly also for computational resources. Curation never sleeps. One might question the cost in complexity (engineering) or computational resources (curation), but this question cannot be answered without taking into account the competitive advantage of a well-curated summary. The investment in resources may be justified to maintain a better model of the past, which might turn out to be the key to good performance of the application.

Scalability requires the cooperation of distinct processes (acquisition, curation, exploitation), possibly well separated over time, and thus these processes must adopt *common conventions*. This too is a major challenge. In exchange, the acquisition process need not worry about conserving space (sampling rate and resolution), as the curation process can offer the illusion of infinite storage. Likewise, the exploitation process can assume that it has access to *all* the past, approximately. Curation relieves the designer of space-related constraints.

From the opportunistic point of view of the exploitation process, the record is a resource to be mined. It must make (or make use of) two abstractions. First, the same value of a statistic might reflect very different values of the data. Second, each set of data can be summarized by different statistics (or the same statistics with different parameters or alignment). The need to make these abstractions, rather than access data directly, is a major challenge.

### Is it worth the trouble?

Given the complexities and challenges, it is tempting to dismiss all talk of statistics, curation, or abstraction, and invest effort instead in storing as much "real" data as possible. Naively, it seems silly to squander space on statistics. And yet, the situations are numerous where precious data are *completely* lost, either because they were too old, or because we were reluctant to discard old data to make room for new. It would be nice at least to have a record of what was lost. And, space permitting, perhaps a few statistics to describe the lost data in more detail? What was new yesterday is old today, and tomorrow will be older still, and yet even the oldest information might someday prove relevant.

A recommendation from this paper might be that *all* data storage should be scalable. Data should be designed to work in five, twenty, or a hundred years, whatever the accrual rate or future costs and constraints. Abundant storage to conserve all data in full is ideal, but failing that, the system should be able to fall back on a strategy that minimizes the impact of data loss. Continued curation throughout all time is ideal but, failing that, the data should at least be in a format that allows a future curation system to resume the task. Looking around, it does not seem that *any* data or systems yet meet this recommendation.

### What is new

Exponential increases in data size and accrual rate challenge many fields, including science, social media, finance, and commerce (Cisco 2017; Feigelson et al 2012; Hilbert and Lopez 2012; Lichtman et al 2014; Stephens et al 2015; Gillings et al 2016). De facto, many data are lost, either because they could not be recorded, or because their record was deleted for lack of space. There are two approaches to avoid loss of precious data. One is to invest in more storage and compute power, the difficulty being to harness them effectively at larger scales. The word *scalability* almost invariably refers to hardware or software solutions to accommodate larger data or more massive data streams.

The other approach is to design more compact summary representations so as to have less to store. A challenge then is to ensure that the summary too remains sufficiently compact in the limit of large data: its size increases in size exponentially unless the scale ratio also increases. Merely applying a larger ratio to the next batch of data is not satisfactory, as it would lead to better resolution for old data than new, the opposite of what is needed (Idea 11). Instead, a new ratio must be applied retroactively to the *already stored summary*, which is what this paper is about.

Many of the ideas in this paper can be found in some form or another in the literature on statistics, data science, or machine learning (James et al 2014, Donoho 2015, Schmidhuber 2014, Leskovec et al 2020) in particular the literature on massive *data streams* (e.g. Aggarwal 2007, 2013, Cormode et al 2012, Bifet et al 2018, Cormode and Yi 2020, Leskovec et al 2020). Concepts of *summary*, *indexing* and *search* are ubiquitous, as is the use of statistics for that purpose. The ability to *aggregate* those statistics, exactly (e.g. mean or min/max) or approximately (e.g. quantiles or heavy-hitters) is often discussed in the context of queries (e.g. Yi et al 2014). There is less attention to what Agarwal et al (2013) call *mergeability* (similar to what we call scalability). However, the idea of summarizing a stream of summaries, repeatedly and ad infinitum, is rarely put forward. A scalable index was introduced into the MPEG 7 standard (ISO 2001, de Cheveigné 2002) and at least two patents describe scalable statistical summarization schemes (de Cheveigné 2001, Agrawal and Vulmini 2017).

The distinction between an exploitation process that knows the value of the data, and a curation process that does not, is rarely explicit, although it arises every time data are reused for a different purpose, or a decision must be made as to the choice of a sampling rate, or whether to discard data. The idea of storing a summary together with a program to curate it appears in the definition of a *sketch* (Bifet et al 2018). The concept of *curation as a service* to answer queries about the summary is the essence of a database manager.

### Learning

The past is useful to predict future *observations* (predictive coding, e.g. Friston 2018) or the future *state of the world*, or the best future *actions* to perform (e.g. Tishby and Polani 2011, Schmidhuber 2019), goals that are related but not identical. A predictive model is fit to past observations (learning), and then applied on an ongoing basis to subsequent observations. A large number of observations of the world, including interactions with it, are thus "distilled" into parameters (weights) and state variables (activations) of the model. Those can be construed as a form of record or "memory" of the past, but they are typically much fewer than the observations, so this is a highly reduced memory. It may nonetheless be entirely sufficient for the needs for which the model was trained.

However, a new need might arise that requires access to the past, *including the past prior to the advent of this new need*. If observations before that moment have been distilled for the old purpose and then deleted, they are no longer available to learn a new model. Records optimized for the needs of the old application might not be helpful for the new. Memory should also cater for those *unseen* needs, and one might actually use this as a definition of "memory" as distinct from the trace that results from learning. Ideally, every bit of the past should be conserved (Schmidhuber 2009), a goal that is unreachable for lack of space but worth approximating, which is what this paper is about.

In a typical learning algorithm, parameters are adjusted incrementally and thus old data are diluted by new data, making it hard to capture dependencies over a long (or variable) time interval. The success of the Long Short Term Memory (LSTM) (Hochreiter and Schmidhuber 1997) has been ascribed to its ability to "hold" a datum over many time steps, and more recently there has been a flurry of proposals that associate a neural network with an external memory (e.g. Sukhbaatar et al 2015, Weston et al 2015, Graves et al 2016, Kaiser et al 2017, Ritter et al 2018, Goyal et al 2022) with a similar motivation.

In some of these proposals, the memory is read/write, serving as a sort of buffer or "working memory" (Graves et al 2016), in others it is closer to a read-only memory that stores actual records of past episodes, possibly via some task-dependent embedding (e.g. Autume et al 2019, Goyal et al 2022). Large performance improvements have been attributed to adding an external memory to a learning algorithm. The perspective developed in this paper might help to expand some of these ideas.

One might wonder if *compression* could meet the goal of fitting an infinite past into finite storage, using statistical regularities to remove unnecessary redundancies (Barlow 1989, Schmidhuber 1992, Friston 2018), in place of summarization. A fixed compression factor won't do, obviously, but perhaps a sustained rate of discovery of *new* rules to achieve better compression might somehow keep up? It has been suggested that we (humans) are hardwired to constantly search for such rules (Schmidhuber 2009). There is little reason to expect a large reduction in size for lossless compression, but lossy compression might have greater success. However, it is then unclear how the optimally compressed code for $[x(i), i = -\infty \ldots t]$ could be derived by merging $x(t)$ with the optimally compressed code for $[x(i), i = -\infty \ldots t-1]$.

Summarization constrains a learning model to make use of a particular subspace of the data, thus inducing a form of inductive bias (Botvnick et al 2019). Dimensionality reduction per se, forced by space limits, may be beneficial to prevent overfitting, but performance will depend on whether the chosen subspace includes the relevant dimensions. The curation heuristics hopefully make this more likely. They do so on the basis of a priori logical or physical arguments, analogous to weight-tying in convolutional neural networks, rather than via learning. However, the *balance* between heuristics is a plausible target for "slow learning" (Botvnick et al 2019). This requires the curation process to adjust its program on the basis of direct instructions from the exploitation process, or indirectly from the history of queries, actions and rewards. In contrast to standard learning algorithms, here the tuning operates merely on the choice of statistic to use in the future to rescale the statistical summary to fit storage.

### Heuristics

Physical laws invoked by Schmidhuber (1992) as a basis for statistical regularity are to some extent also captured by the curation heuristics mentioned earlier. Physical objects have inertia, processes have a temporal extent, consequential events tend to have (by definition) consequences as well as causes, and these might show up as regularities within the stream of observations. Regularities can be exploited as heuristics to guide summarization, and they may also serve as features, the heuristics making it more likely that those features survive the curation process. A regularity is of interest to predict future parts of the stimulus, and because its applicability at a certain time (but not others) is noteworthy. Barlow (1990) argued that the statistics of *all* aspects of the data should be recorded, even if *not* noteworthy, because knowledge of their distribution is a precondition for the detection of significant associations.

As mentioned earlier, incremental adjustment of a parameter implies an exponential down-weighting of remote values, diluted by new. The recency heuristic, enforced by regular rescaling, also implies an exponential down-weighting of the past, a difference being that its effect is to *thin* dimensions, rather than decay a value. Individual values may be merged with their neighbours (i.e. diluted) in the rescaling process, but as long as the record retains the structure of a time series, remote events can be distinguished from new. A similar effect can, in principle, be obtained by maintaining a basis of integrators with exponentially-shaped windows, from which a window of arbitrary shape can be synthesized (following the principle of the Laplace transform, Howard 2018), but that is a rather roundabout way to select a datum remote in time.

The *non-stationarity heuristic* prioritizes merging data so that interval boundaries tend to coincide with changes in statistics (changepoints). There is an abundant literature on changepoint detection within data streams (e.g. Aggarwal 2013), and more generally segmentation and clustering. Depending on the constraints, changepoint detection may occur offline (with access to all time) or online (with access only to the past). Rescaling as described here is intermediate between the two: merging can be deferred during the curation process, but not for too long, as there is pressure to make room for new data. Rescaling is analogous to agglomerative clustering (as opposed to divisive), and the non-stationarity and maximal KL-divergence heuristics are related to probabilistic changepoint detection (e.g. Adams and MacKay 2007).

In an online setting, a change in statistics is referred to as *surprise* (or surprisal) (Baldi and Itti 2010, Gershman et al 2014): new data that fit the distribution of previous data are unsurprising and do not warrant a distinct statistical description. The offline setting allows more subtlety: a distinct description might also be warranted if the *old distribution does not fit that of the new data*, as KL divergence is not commutative. The online setting requires defining a threshold to decide whether to set a changepoint, with the risk that a permissive threshold might lead to too many samples to fit storage. The offline setting requires no such threshold: the least distinct intervals are merged until the statistics fit the storage.

The statistical summary is an *abstraction* of the data, given that different data sets could yield the same summary, and a *description* to the extent that this abstraction captures a noteworthy aspect of the data, for example its texture. As argued earlier, in the limit of large data, (a) small scale details are too numerous to be all worthy of attention, and (b) storage constraints prevent them being retained. Summary statistics are known to be perceptually relevant in vision (Ariely 2001, Alvarez 2011, Whitney and Leib 2018) and audition (McDermott et al. 2013, McWalter and Dau 2017, McWalter and McDermott 2018). As data become larger, those statistics too require summarization and abstraction, ad infinitum in the limit of large data. Constructs such as the scattering transform (Andén and Mallat 2014) or modulation cascade process (Turner and Sahani, 2008) handle this limit uniformly, using the same operations at each scale.

## Conclusion

Data are rapidly growing in size and importance for technology and society, but the boundless accumulation of new data poses a difficult challenge. This challenge is addressed here by postulating that data must be stored in a statistical format that allows them to be rescaled. The passive process of storage is replaced by the active process of curation.
Its function is to maximize the future value of the data given limited storage. This requires complex data structures and sophisticated computation, an inevitable consequence of the need to accommodate boundless data within bounded space while maximizing their value for future purposes.

## Acknowledgments
This essay benefited from comments by Eli Nelken and Malcolm Slaney, as well as earlier discussions with Tali Tishby. This work was supported by grants ANR-10-LABX-0087 IEC, ANR-10-IDEX-0001-02 PSL, and ANR-17-EURE-0017

# Appendix A - Scalable statistics

Basic statistics such as mean, variance, extrema, cardinality, histogram, and for multivariate data, covariance and convex hull are described in the text. All those statistics are scalable. Applied repeatedly to a stream of data, they yield a time series of statistics (values). Slow fluctuations in the data are preserved in this time series, but faster fluctuations are lost. This appendix lists a few additional statistics (the list is not exhaustive).

Information about faster fluctuations can be preserved, in part, by the short-term *power spectrum*. With an appropriate definition, this statistic is both concise, relative to the raw data, and scalable. A plausible choice is a wavelet-based spectrum with a logarithmic frequency axis, an example of which is described in Appendix C. Applied repeatedly, the statistic results in a time series of spectrum coefficients. As data accumulate, it may be useful to summarize the fluctuations of these coefficients as a *second-order power spectrum* (also called *modulation spectrum*). The modulation spectrum describes the temporal pattern of fluctuation of power spectrum coefficients. Its coefficients can be summarized as a *third-order power spectrum*, and so-on, a hierarchical analysis scheme that has been formalized as the *scattering transform* (Anden and Mallat, 2014; Bruna and Mallat, 2013). With appropriate definitions, all of these statistics are scalable.

It may be useful to keep tabs on the covariation of parallel data streams, for example to spot a causal relation between them, or a hidden cause that affects them both. A statistic for this purpose is the *correlation coefficient*, sum over time of the cross-product of two time series. Correlation is scalable, as is the *cross-correlation function* (correlation calculated for a range of relative delays between streams). Multiple and/or multivariate streams can be described by a *covariance matrix*, which in turn allows PCA (principal component analysis) to be applied to factor out redundancies and reduce the cost of tracking correlation. With care, all these tools can be deployed in a scalable manner.

Some data samples might be less relevant than others, for example due to glitches (outliers). To avoid contaminating the statistics, the data stream can be augmented with a time series of *weights* to be applied in the calculation of the statistics and their rescaling. The weights also document the fact that some data samples were ignored or down-weighted. Weights are scalable.

Cardinality and histogram are applicable also to non-numeric data. Non-numeric items can be represented as pointers or indices into a *dictionary*, groups of such items can be summarized as a histogram, each bin labelled by a dictionary entry (Fig. A1), and a stream of data can be summarized as a time series of such histograms, the dictionary being shared across histograms. This representation is scalable: histograms are merged by summing corresponding bins and, if necessary, merging dictionaries.

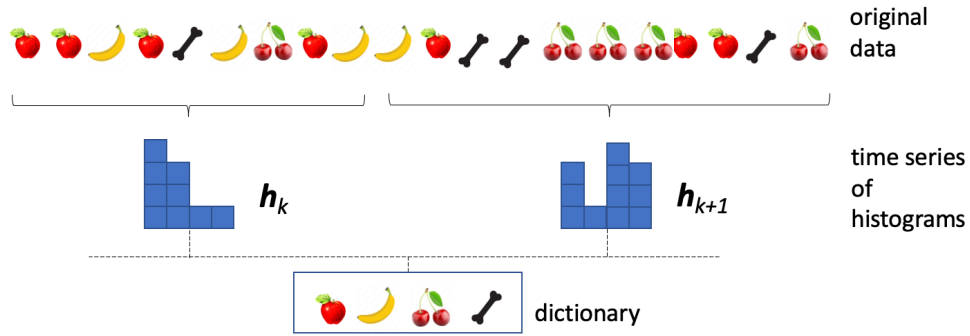

Fig. A1. A stream of arbitrary data can be summarized by a time series of histograms with bins defined by a dictionary shared across histograms. The representation can be rescaled by summing histograms of consecutive intervals, and it is also possible to simplify the dictionary (e.g. merge fruits into a single category) to save space.

The dictionary can be created and updated on the fly: an item that finds no match can be used to create a new entry. If the dictionary becomes too large, its least common entries can be dropped (e.g. Agarwal et al 2013) or similar entries can be merged, counts of deleted entries can be transferred to an "outlier" bin, and so on.

A dictionary can be used to implement an "episodic" memory, by storing recurrent patterns as dictionary entries and comparing incoming data to these entries (Fig. A2), as in vector quantization (Gray 1984).

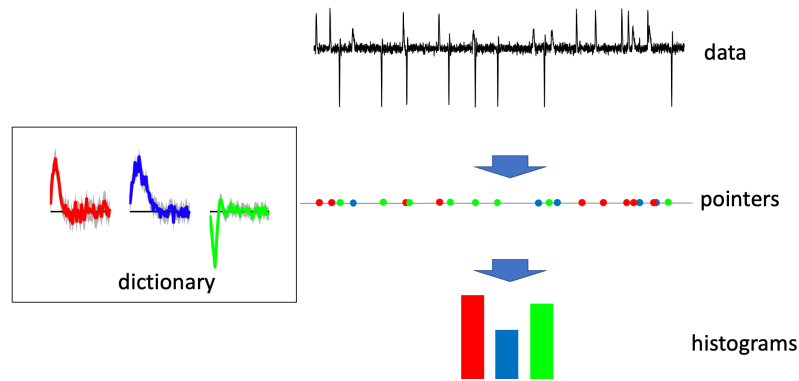

Fig. A2. An example of dictionary-based episodic memory. Stereotypic events in the input stream (top) are represented by their shape and positions in time, which can be summarized by a time series of histograms (bottom). Here, each dictionary entry is defined by a mean (colored line) and standard deviation (gray band), estimated empirically from the data.

In the real world, items are never exactly identical, so the item-to-entry mapping must have some slop. It may then be useful to keep track of the *distribution* of the items that mapped to each bin. This distribution, once learned, can then also serve to define the entry (Fig. A3). The distribution can be parametrized by statistics managed in a scalable fashion, allowing a trade-off between accuracy of representation of the distribution and cost. The statistical summary of the data then consists of both the dictionary and the bins, both scalable and both undergoing an active process of curation.

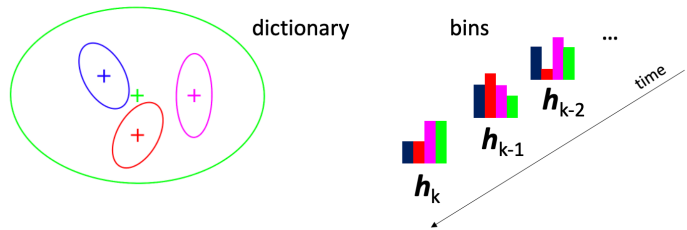

Fig. A3. Left: the dictionary consists of distributions. Each new data sample is assigned to the closest distribution (Mahalanobis distance), and the corresponding bin is incremented. The three narrow distributions model clusters in the data, the wider distribution (green) catches outliers.

The scheme represented in Fig. A3 needs to be bootstrapped. One strategy is to create entries promiscuously, space permitting, and then, when space becomes tight, merge those that are similar and delete those that occur rarely (Fig. A4) as in the BFR algorithm (Leskovec et al 2020). This strategy is all the more successful as there is more storage to accommodate all the "wannabe" patterns (Ackerman and Dasgupta 2014), which justifies the assertion "more memory allows better memory" (Idea 21).

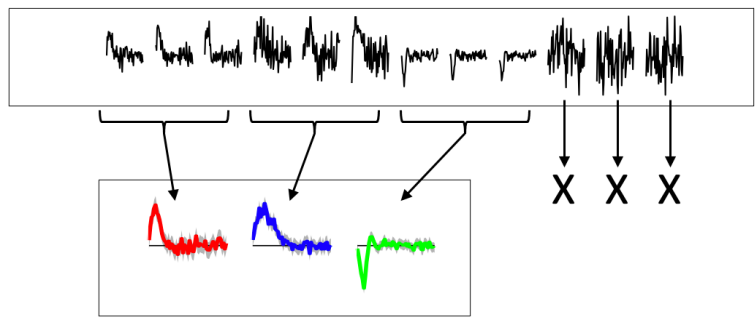

Fig. A4. Bootstrapping the dictionary. As many patterns (snippets) are recorded in the dictionary as space permits (top). When space becomes scare, patterns that are similar are merged (bottom) and those that recur rarely are deleted (right).

Data *selection*, including *down-sampling*, fits the definition of statistic and is scalable, as does *deletion* as a limit case. *Random selection* and *random projection* are scalable in a probabilistic sense. Other summary types such as linear sketches, locality sensitive hashing, Bloom filters, etc. may be scalable or approximately scalable (e.g. quantiles or quantization). What counts is whether they remain useful in the limit of large data.

## Appendix B - Heuristics

Heuristics guide the choice of statistics on the basis of logical or physical constraints or past experience with similar data or similar applications. Applying them is expected to give better results across a range of future applications, albeit not *optimal* for any particular goal.

### Recency

Rescaling implements the heuristic that recent observations are more valuable than old (Idea 11). For example, if every part of the record is rescaled by the same factor, the space devoted to each observation decreases exponentially in time.

This heuristic takes care of itself, but one might want to further ensure that storage is apportioned more or less equally across scales. For example, given $M$ bits of available storage and $N$ bits of data to summarize, one can apportion $M/log2(N/M)$ bits to each level of a binary rescaling scheme. In other words, $M/log2(N/M)$ bits are assigned to full-resolution data, the same to data summarized by a factor 2, and so on. If a sound recorder can record 24 hours of sound at full resolution, multiplying its storage by a factor of 9 would cover a year, and multiplying that again by a factor of 5 could have covered a span from the Big Bang to the present. With the same storage one could instead accommodate 45 days at full resolution, but then one would be blind (or deaf) to everything that happened earlier...

Note that this differs from the exponential decay of the state of a linear system or recurrent neural network. Thinning by repeated rescaling reduces the *number of dimensions* attributed to a set of data, not just the weight of their contribution to a single dimension.

### Slowness

Meaningful processes tend to extend over time, and the inertia of physical objects prevents erratic movements, suggesting that useful dimensions may be *slow*. This is the rationale for an analysis technique known as *slow feature analysis* (Wiskott and Sejnowski). Dimensions that fluctuate erratically (or with statistics that fluctuate erratically) might be prioritized for deletion.

### Non-stationarity, surprisal

In counterpoint to the previous heuristic, processes of interest are often non-stationary, and a non-stationarity might signal an event of interest. Details of the non-stationary part may deserve more bits than the intervening quiescent intervals. This heuristic might be implemented by finding change points indicating a change in data distribution, for which there is an abundant literature (Adams and MacKay 2007, Aggarwal 2007, 2013).

As an example, $S_{k-1}$ and $S_k$ being distributions parametrized by consecutive samples of the summary statistic, if the KL divergences $d(S_{k-1} \| S_k)$ and $d(S_k \| S_{k-1})$ are both small, then the underlying data are similarly distributed (from the viewpoint of this statistic) and there is little merit in describing both statistics. Scanning the summary, merging adjacent samples with small KL divergences, would implement the heuristic. As an aside, merging *non-consecutive* samples (replacing each by a pointer to the same merged sample) would result in a non-sequential structure analogous to the dictionary-based episodic memory mentioned earlier.

### Maximal KL divergence

Whereas the previous heuristic (non-stationarity) focuses on time (merging intervals with small KL-divergence), this heuristic focuses on statistics (deleting statistics that yield small KL-divergences). KL divergence is all the more useful as it can take large values, as these allow for reliable pruning. This depends of course on the data, but also on the choice of statistic used to summarize them. For a given dataset, not all statistics are equal in this respect. Statistics that tend to yield small values of KL divergence between different subsets of the data should be prioritized for deletion. It is not useful to label separately data that can't be distinguished from the labels.

### Joint diagonalization
A problem for distance measures applied to multidimensional data is that the distance depends on how the dimensions are weighted, which affects for example our ability to cluster the data. Divergence is not immune to this problem, as data in high dimensions are sparse, making "true" distributions hard to estimate from data. A partial solution is to reduce dimensions to those that are most "relevant" according to some criterion. A general tool for this purpose is *joint diagonalization* (Särelä and Valpola 2005; Valpola 2005; de Cheveigné and Parra 2014), which conveniently relies on covariance matrices, a scalable statistic. Joint diagonalization can be understood as a kind of PCA in which variance is replaced by a measure of relevance. As an example, if two subsets of data are labelled with their covariance matrices, joint diagonalization of those matrices produces a transform matrix that can be used to select the subspace within which they differ most. Distributions can be more reliably estimated within that subspace because the data are less sparse than in a larger space, so KL divergence can more effectively decide if those subsets should be prioritized for merging.

### Other cues to relevance
*Prior access* to some part of the record is a hint to its relevance (particularly if that access was followed by action or reward). A pattern that *recurs* may be worth keeping longer, as may be one that is *predictive* of another, or is *predicted* by another, or merely *cooccurs* with another pattern, suggesting a common cause. The other pattern could belong to the same data stream, or a different stream, or it could consist of an action, or a reward. The rationale is that data that are marked in this way are unlikely random, and should thus be given some reprieve in the drive to rescale. The amount of reprieve to give in each case is a likely target for optimization based on past experience. Additionally, there may be domain-specific *invariances* in the data as exploited by CNNs.

### Competition between heuristics
Different heuristics suggest different priorities, and the outcome of summarization (and value of the summary for future use) depends on how they are ranked or combined. A default strategy might be to select a few dimensions favoured by each heuristic and include them all in the summary. This strategy is made easier if there is more storage space, another example of "more memory leads to better memory". Alternatively, the choice of statistics can be refined based on feedback from prior use of similar data, in a process akin to (but distinct from) model learning. The distinction from model learning is that it involves a coarse-grained choice between heuristics rather fine-grained tuning of weights.

## Appendix C - Binary scaled summary
This section describes one possible statistical summarization scheme. The input stream is a time series $x(t)$, where $t$ is an ever-increasing time index, summarized by a statistical record $S(t)$ continuously updated by merging samples two-by-two.

The record $S(t)$ consists of a time series of summary statistics calculated from past values of the input data according to a binary scheme. Specifically, statistic $s_k(t/2^k)$ is the result of applying the statistic to pairs of data samples, then repeatedly merging those samples by pairs to obtain a single value summarizing an interval of size $2^k$ as in Fig. C1.

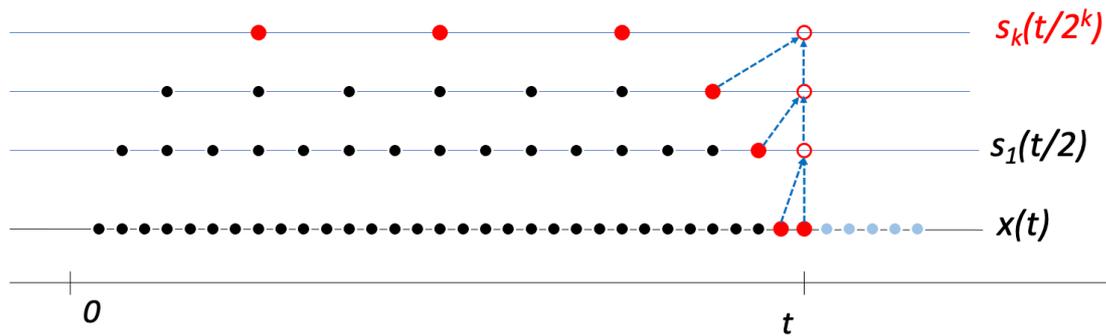

Fig. C1. Binary scaled summary. Incoming samples of $x(t)$ are grouped in pairs and summarized to form samples of $s_1(t/2)$. These in turn are paired and summarized repeatedly to obtain a time series $s_k(t/2^k)$ that summarizes the data at a rate $2^k$ times slower than the original. The summary $S(t)$ consists of the samples labelled in red. This summary is updated at every time index (blue arrows).

Past samples of $x(t)$ may be discarded to save space, as well as past samples of $s_{k'}(t/2^{k'})$, $k' < k$, leaving only the samples labelled in red in Fig. C1. The summary record $S(t)$ then consists of these samples, updated at every time index (blue arrows).

In this example, all available storage is devoted to one scale, $k$. One might instead want to allocate storage evenly across scales, so as to assign uniform importance across time on a logarithmic scale, as illustrated in Fig. C2. If recording started a time $t = 0$ we have $K = log2(t)$ scales to accommodate. Given $N$ storage slots we can assign $N/K$ samples to each.

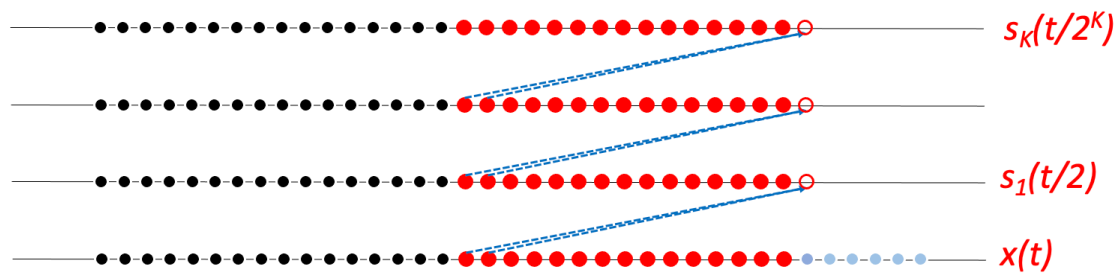

Fig. C2. Binary scaled summary with equal storage allocated to each scale. Samples in black are discarded after summarization (blue arrows). The time scales on different lines are different, but for each the rightmost red point represents 'now'. The summary record $S(t)$ consists of all samples in red.

The binary summarization scheme of Fig. C2 supposes uniform sampling at each scale. Non-uniform sampling is also possible by deleting unwanted samples, but care must be taken to ensure scalability: statistics that depend on deleted samples must be calculated before deletion.

Some simple statistics are:
- *Cardinality* $n_k = 2^k$, represented implicitly.
- *Mean*, calculated recursively as $m_{k+1}(j/2) = [m_k(j) + m_k(j-1)]/2$, where index $j = t/2^k$.
- *Maximum* and *minimum*, similarly calculated.

- *Histogram* bin counts, calculated recursively $h_{k+1}(j/2, i) = h_k(j, i) + h_k(j-1, i)$ where $i$ is the histogram bin index.
- *Variance*, calculated as
$$v_{k+1}(j/2) = [v_k(j) + v_k(j-1)]/2 + [(m_k(j) - (m_{k+1}(j/2))^2 + ((m_k(j-1) - (m_{k+1}(j/2))^2]/2.$$

The formula for variance involves two terms: one aggregates the variance from the preceding level (similar to the formula for mean), and the second augments it with variance carried by the means at the preceding level. Variance thus 'grows' by a non-negative amount at each rescaling. Building on this idea, one can decompose the variance over an interval of size $2^K$ as a sum of non-negative terms:

$$v_K(j) = \sum_{k=1}^{K} v_K^k(j) \quad\quad\quad (A1)$$

where $j = t/2^K$. Each term $v_K^k(j)$ can be understood as representing the portion of variance associated with scale $k$. The vector of values $v_K^k(j), k = 1, \cdots K$ is referred to here as the "*scale-wise variance*" of the interval of data. Since the variance can also be expressed as a sum of squares of individual data samples we have:

$$\sum_{i=1}^{2^k} x(t-i+1)^2/2^K = \sum_{k=1}^{K} v_K^k(j) + m_K^2(j)/2^K \quad\quad\quad (A2)$$

which is similar to Parseval's relation. This suggests that one can treat scale-wise variance as a form of power spectrum with a log frequency axis (inverted because lower values of $k$ represent more rapid fluctuations).

Scale-wise variance, associated with mean, is scalable: terms $k = 1:K$ of $v_{K+1}^k(j/2)$ are calculated by averaging consecutive terms of $v_K^k(j)$, while $v_{K+1}^K(j/2)$ is derived from consecutive terms of $m_K(j)$. The scale-wise variance vector thus grows by one sample at each rescaling. Like a spectrum, scale-wise variance describes the pattern of fluctuation within the interval. It also gives an indication as to 'lumpiness' of the distribution of values, which may be useful for search (a lumpier distribution promises more rapid pruning).

Applied to successive intervals of size $2^K$, each term of the scale-wise variance forms a time series that itself can be summarized using statistics such as the mean, variance, or scale-wise variance. Such 'summarization of summarizations' can be repeated to obtain a hierarchical representation akin to the scattering transform of Andèn and Mallat (2014). The first order transform (scale-wise variance of data samples) corresponds to a spectrum, the second order transform (scale-wise variance of time series of scale-wise variance coefficients) and so on. Higher-order transforms capture longer term fluctuations.

A practical detail is worth considering. Scale-wise variance (like variance) is quadratic in the amplitude of data samples, second-order scale-wise variance is quartic, and so-on. This may cause these statistics to be over-sensitive to high-amplitude values. To address this, a compressive transform (e.g. square root, or cubic root) can be applied to each time series of coefficients before handing it over to the next stage. This obviously affects the semantics of the summary.

Scalewise variance, and its higher-order versions, are more complex that basic statistics such as variance, but they address an unavoidable need: discover and represent meaningful abstract patterns at the *largest time scale*. Those higher-order transforms are expected to become more useful in the limit of larger data.